\documentclass[journal]{IEEEtran}
\ifCLASSINFOpdf
   \usepackage[pdftex]{graphicx}
\else
\fi
 \usepackage{multirow}

\usepackage[cmex10]{amsmath}
\usepackage{amssymb}
\usepackage{algorithm,algorithmic}
\begin{document}
%
\title{Less-forgetting Learning in Deep Neural Networks}
%
%
%

\author{Heechul~Jung,~\IEEEmembership{Member,~IEEE},
Jeongwoo~Ju,~\IEEEmembership{Student Member,~IEEE},\\
Minju~Jung,
        and~Junmo~Kim,~\IEEEmembership{Member,~IEEE}
\thanks{Heechul Jung, Minju Jung and Junmo Kim are with the School of Electrical Engineering, Korea Advanced Institute of Science and Technology, Daejeon, Republic of Korea, e-mail: \{heechul, alswn0925, junmo.kim\}@kaist.ac.kr. Jungwoo Ju is with the Division of Future Vehicle, Korea Advanced Institute of Science and Technology, Daejeon, Republic of Korea, e-mail: veryju@kaist.ac.kr}
}

%
%

\markboth{Journal of \LaTeX\ Class Files,~Vol.~xx, No.~x, xxxx~xxxx}%
{Shell \MakeLowercase{\textit{et al.}}: Bare Demo of IEEEtran.cls for Journals}
%



\maketitle

\begin{abstract}
A catastrophic forgetting problem makes deep neural networks forget the previously learned information, when learning data collected in new environments, such as by different sensors or in different light conditions.
This paper presents a new method for alleviating the catastrophic forgetting problem. Unlike previous research, our method does not use any information from the source domain. Surprisingly, our method is very effective to forget less of the information in the source domain, and we show the effectiveness of our method using several experiments. Furthermore, we observed that the forgetting problem occurs between mini-batches when performing general training processes using stochastic gradient descent methods, and this problem is one of the factors that degrades generalization performance of the network. We also try to solve this problem using the proposed method. Finally, we show our less-forgetting learning method is also helpful to improve the performance of deep neural networks in terms of recognition rates.
\end{abstract}


\IEEEpeerreviewmaketitle

\section{Introduction}
\label{sec:int}
\IEEEPARstart{D}{eep} neural networks (DNNs) have grown to nearly human levels of recognition in identifying objects, faces, and speeches \cite{taigman2014deepface, graves2013speech, szegedy2014going, simonyan2014very}. Despite this advancement of deep learning, remaining issues still exist; a catastrophic forgetting problem is the one of these remaining issues \cite{goodfellow2013empirical}.
The problem is an important issue in DNNs since this enables an improvement in the performance of DNNs in several important applications such as domain adaptation and incremental learning.

A catastrophic forgetting phenomenon is often observed when performing domain adaptation because the distribution of source data is significantly different from the distribution of target data. 
For example, consider the traditional transfer learning protocol (weight copy $\rightarrow$ fine-tuning) for adapting to a new domain. Usually, the network pre-trained using the original data (source domain) is used as initial weights for adapting to the new data (target domain) \cite{patel2015visual, pan2010survey}.
During learning new data from the target domain, it is natural that the network forgets the previously learned information from the source domain. Even if source and target domains are nearly homogeneous, the network forgets the information about source data.

Several researches have been performed for alleviating such problem. 
Srivastava et al. proposed a local winner-take-all (LWTA) activation function that helps to prevent the catastrophic forgetting \cite{srivastava2013compete}. This activation function has the effectiveness of implicit long-term memory. 
Recently, several experiments of a catastrophic forgetting problem in DNNs were empirically performed in \cite{goodfellow2013empirical}. The paper shows a dropout method \cite{Hinton2012, srivastava2014dropout} with a Maxout \cite{goodfellow2013maxout} activation function is helpful for forgetting less of the learned information.
However, these kinds of methods are not explicitly used for the unforgetting of previously learned information, which means that it does not guarantee the unforgetting ability.

An unsupervised approach was proposed in \cite{goodrich2014unsupervised}. Goodrich et. al extended this method to a recurrent neural network \cite{goodrichmitigating2015}.
These methods compute cluster centroids while learning the training data in source domain. They also use the computed centroids for the target domain. Consequently, these are not generally applicable for the pre-trained model, and these are not adequate for our problem settings. \cite{goodrich2014neuron} and \cite{lancewicki2015sequential} have similar issues mentioned above.

\begin{figure}[t!]
\begin{center}
\includegraphics[width=4.5cm]{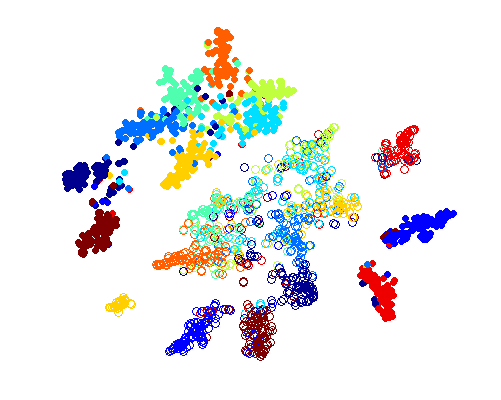}~~~~
\includegraphics[width=4.5cm]{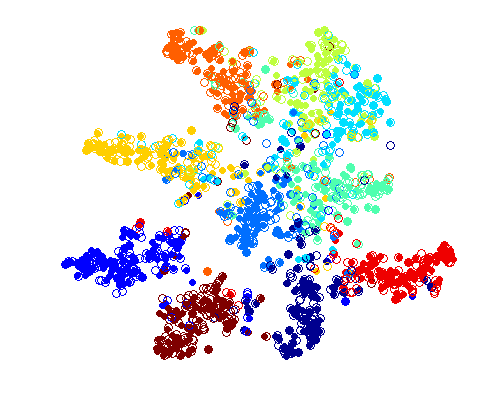}\\
\small
(a)~~~~~~~~~~~~~~~~~~~~~~~~~~~~~~~~~~~~~~~~(b)
\caption{\textbf{Visualization of the feature space for ten classes using t-SNE \cite{van2008visualizing}.} Each color represents each class. Filled circles denote features of the source training data extracted by the source network. Circles represent features of the source training data extracted by the target network. (a) Traditional transfer learning method. (b) Proposed method.}
\label{fig:feature}
\end{center}
\vspace{-5mm}
\end{figure}

In this paper, we try to solve a catastrophic forgetting problem in DNNs by using a new learning method.
The proposed method has the ability to maintain the original feature space of the source domain, even if the network does not see any previous training data. Therefore, the method forgets less of the information obtained from the source data compared to the traditional transfer learning method. Figure \ref{fig:feature} shows the feature spaces of the traditional transfer learning method and the proposed method. In the proposed method, features of the same source class and target data are well clustered, even if re-training only using the target data is finished. 

Our proposed method is also applicable to the learning method from scratch, which is a general training protocol using stochastic gradient descent methods. We observed that the forgetting problem also arises between mini-batches because mini-batches are small datasets subsampled from a large amount of whole data. We also deal with such problems.
We summarize our main contributions as follows:
\begin{itemize}
\item We propose a less-forgetting learning method to alleviate a catastrophic forgetting problem in DNNs.
\item We observe that a catastrophic forgetting problem also occurs between mini-batches when using a stochastic gradient descent learning.
\item Furthermore, we show our less-forgetting learning is effective to solve the problem, and it gives better generalization performance.
\end{itemize}


\subsection{Less-forgetting Problem}
\label{sec:prob}
To theoretically show the less-forgetting problem, we assume that we have the weight parameters of the pre-trained model for the source domain, and training data are given for the target domain (target data). Consequently, data for the source domain (source data) are not accessible.
For more clarity, we explain the problem using mathematical expressions as follows.

We denote that the dataset for the source domain is $\mathrm{D}^{(s)} = \{(x_i^{(s)},y_i^{(s)})\}_{i=1}^{n_s}$, and the dataset for the target domain is $\mathrm{D}^{(t)} = \{(x_i^{(t)},y_i^{(t)})\}_{i=1}^{n_t}$, where $n_s$ and $n_t$ are the number of datasets of the source domain and target domain, respectively. Furthermore, $x_i^{(\cdot)}$ is the training data, and  $y_i^{(\cdot)}$ is the corresponding label. These two datasets are mutually exclusive, and each dataset has both the training and validation datasets as follows:
$\mathrm{D}^{(s)} = \mathrm{D}_t^{(s)} \cup \mathrm{D}_v^{(s)},
\mathrm{D}_t^{(s)} \cap \mathrm{D}_v^{(s)}=  \varnothing, \mathrm{D}^{(t)} = \mathrm{D}_t^{(t)} \cup \mathrm{D}_v^{(t)}$ and $\mathrm{D}_t^{(t)} \cap \mathrm{D}_v^{(t)}=  \varnothing, $
where $\mathrm{D}_t^{(\cdot)}$ and $\mathrm{D}_v^{(\cdot)}$ are training and validation datasets, respectively.

The source network $\mathrm{F}(x; \theta^{(s)})$ for the source domain is trained using $\mathrm{D}_t^{(s)}$, where $\theta^{(s)}$ is a weight parameter set. The initial values of the weights are initialized randomly with $\mathcal{N}(0,\sigma^2)$, a normal distribution.
The trained weight parameters $\theta^{(s)}$ for the source domain are obtained by using the dataset $\mathrm{D}_t^{(s)}$.
The target network $\mathrm{F}(x; \theta^{(t)})$ for the target domain is trained using the dataset $\mathrm{D}_t^{(t)}$. 
Finally, we obtain the updated weight parameters $\theta^{(t)}$.
In order to satisfy the less-forgetting condition, $\mathrm{F}(x;\theta^{(t)}) \approx \mathrm{F}(x;\theta^{(s)})$ for $x \in \mathrm{D}^{(s)}$.

\section{Less-forgetting Learning}
\label{sec:proposed}
In DNNs, the lower layer is considered as a feature extractor, and the top layer is regarded as a linear classifier.
This means that the weights of the softmax function represent a decision boundary for classifying the features.
Due to the linear classifier on the top layer, the features extracted from the top hidden layer are usually linearly separable. 
Using this knowledge, we propose a new learning scheme that satisfies the following two properties to reduce the forgetting problem of the information learned from the source domain:\\

\noindent\textbf{Property 1.} \textit{The decision boundaries should be unchanged.}\\\\
\noindent\textbf{Property 2.} \textit{The features extracted from source data by the target network should be present in a position close to the features extracted from source data by the source network.}\\

 
\begin{figure}[t]
\begin{center}
\includegraphics[width=\linewidth]{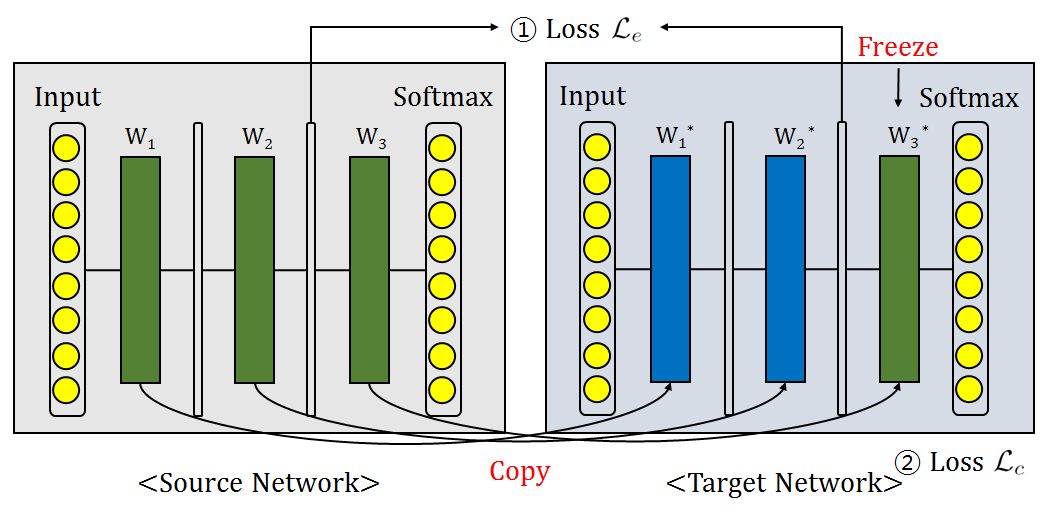}
\caption{\textbf{Conceptual diagram for describing a less-forgetting method.} Our learning method uses the trained weights of the source network as the initial weights of  the target network and minimizes two loss functions simultaneously.}
\label{fig:overview}
\end{center}
\vspace{-3mm}
\end{figure}

\noindent We build a less-forgetting learning algorithm based on two properties. The first property is easily implemented by setting the learning rates of the boundary to zero, but satisfying the second property is not trivial since we cannot access to the source data. Instead of using the source data, we use the target data and show it is helpful to satisfy Property 2. Figure \ref{fig:overview} briefly explains our algorithm, and the details are as follows. 

Initially, like a traditional transfer learning method, we reuse the weights of the source network as the initial weights of the target network. Next, we freeze the weights of the softmax layer to maintain the boundaries of the classifier. Then, we train the network where to simultaneously minimize the total loss function as follows:
\begin{equation}
\mathcal{L}_{t}(x;\theta^{(s)}, \theta^{(t)}) = \lambda_c\mathcal{L}_c(x;\theta^{(t)}) + \lambda_e\mathcal{L}_e(x;\theta^{(s)}, \theta^{(t)}),
\label{eq:total_loss}
\end{equation}
where $\mathcal{L}_t$, $\mathcal{L}_c$, and $\mathcal{L}_e$ are the total, cross-entropy and Euclidean loss functions, respectively. Further, $\lambda_c$ and $\lambda_e$ are the tuning parameters, and $x \in \mathrm{D}^{(t)}$.  Usually, parameter $\lambda_e$ has a smaller value than the value of $\lambda_c$.
The cross-entropy loss $\mathcal{L}_c$ is defined as follows:
\begin{equation}
\mathcal{L}_{c}(x;\theta^{(t)}) = -\sum_{i=1}^{C}t_i \log (o_i(x;\theta^{(t)})),
\end{equation}
where $t_i$ is the $i$-th value of the ground truth label, and $o_i^{(t)}$ is the $i$-th output value of the softmax of the target network. $C$ is the total number of class. In other words, this loss function helps the network to classify the input data $x$ correctly.
In order to satisfy the second property, $\mathcal{L}_e$ is defined as follows:
\begin{equation}
\mathcal{L}_{e}(x;\theta^{(s)}, \theta^{(t)}) = \frac{1}{2} ||\mathbf{f}_{L-1}(x;\theta^{(s)})-\mathbf{f}_{L-1}(x;\theta^{(t)})||^2_2,
\end{equation}
where $L$ is the total number of hidden layers, and $\mathbf{f}_{L-1}$ is a feature vector of layer $L-1$. Using the loss function, the target network learns to extract features which are similar to the features extracted by source network. 

\begin{algorithm}[t]
\footnotesize
\caption{Less-forgetting Learning (LF)}
\textbf{Input:} $\theta^{(s)}, \mathrm{D}_t^{(t)}, N_i, N_b$\\
\textbf{Output:} $\theta^{(t)}$\\
\begin{algorithmic}[1]
  \STATE $\theta^{(t)}\leftarrow\theta^{(s)}$ // initial weights
  \STATE Freeze the weights of the softmax layer.
  \STATE for i=1,$\ldots$,$N_i$ // training iteration
  \STATE ~~~~Select mini-batch set $\mathrm{B}$ from $\mathrm{D}_t^{(t)}$. ($|\mathrm{B}| = N_b$)
  \STATE ~~~~Update $\theta^{(t)}$ using backpropagation with $\mathrm{B}$ to minimize total loss $\mathcal{L}_t(x;\theta^{(s)}, \theta^{(t)})$.
  \STATE end for
  \STATE Return $\theta^{(t)}$.
\end{algorithmic}
\end{algorithm}

 Finally, we build a less-forgetting learning algorithm, as shown in Algorithm 1. $N_i$ and $N_b$ in the algorithm denote the number of iterations and the size of mini-batch.

\section{Less-forgetting for General Learning Cases}
\label{sec:general}

\begin{figure}[b]
\begin{center}
\includegraphics[width=4.5cm]{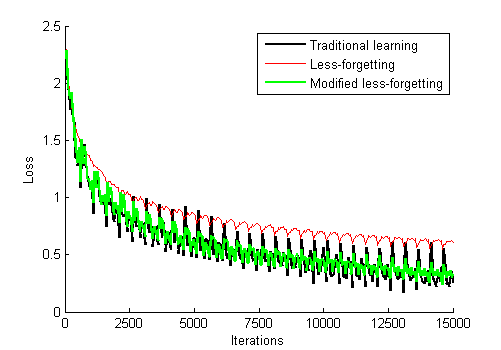}
\includegraphics[width=3.5cm]{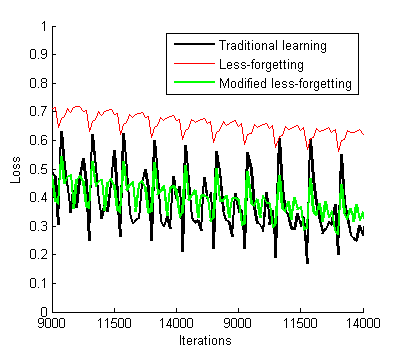}
\caption{\textbf{Graphs for observing forgetting in general learning approach.} The x-axis denotes the value of iteration number, and the y-axis represents the training loss value and training accuracy of a particular batch.}
\label{fig:forget}
\end{center}
\vspace{-5mm}
\end{figure}

It is well known the forgetting problem occurs when learning a new task \cite{goodfellow2013empirical}. In addition, we show that the forgetting problem also occurs when performing general learning process using gradient descent methods. To observe the forgetting phenomenon of the network, we probed the training loss values of a particular mini-batch set, as shown in Figure \ref{fig:forget}. The observing procedure is as follows:
\begin{enumerate}
\item We select a particular mini-batch set $\mathrm{B}$ that will be observed, from the training set $\mathrm{D}_t$. ($|\mathrm{B}|=100$, $|\mathrm{D}_t|=50000$ and $\mathrm{B}\subset\mathrm{D}_t$).
\item Perform a standard stochastic gradient descent-based backpropagation with $\mathrm{D}_t$ and without shuffling on each iteration using three methods such as traditional learning method, less-forgetting method in Algorithm 1 and modified less-forgetting method in Algorithm 2. (total number of epoch = 30)
\item We capture the loss value with $\mathrm{B}$ every 50 iterations (= 0.1 epoch).
\end{enumerate}
Using the experiment, we could identify the forgetting problem when performing traditional learning, as shown in Figure \ref{fig:forget}. It shows periodically oscillating signals, and the one cycle of the signal is exactly the same to one epoch. In other words, the valley is observed at the moment when the network sees $\mathrm{B}$. At the next captured point (after 50 iterations), the loss value was increased. We say that this phenomenon is due to the forgetting problem. In order to satisfy the condition of less-forgetting between mini-batches, the gap between valley and peak points might be narrow.

Figure \ref{fig:forget} shows our less-forgetting method has more smooth graph than traditional learning method. It seems to alleviate the forgetting problem, but the value of training loss is higher than traditional learning method. This is due to the first property in Section \ref{sec:proposed}, and freezing the boundary is an obstacle to learn new data. In the modified less-forgetting algorithm, we often unfreeze the boundary of the network. In addition, we update parameters of the source network using the parameters of the target network. Finally, as shown in Figure \ref{fig:forget}, the green line is more smooth than the black line, and the green line has lower loss values than the red line.

Using this observation, we present a less-forgetting algorithm for general learning cases, as shown in Algorithm \ref{alg:general}. Our algorithm has two main parts (line 3$\sim$5 and line 6$\sim$12). The first part is to switch the parameters of source network which means the network that we want to forget less, and the second part is to unfreeze repeatedly the boundary of the network. If the value of $N_s$ is large, the network does not adapt a new data well. Further, the parameter of $N_f$ plays a role similar to $N_s$. As a result, our algorithm has an ability to forget less the information learned previously. We set the value of $N_s$ is smaller than $N_f$, and we set $N_s=100$ and $N_f=1000$ for all the experiments.

\begin{algorithm}[t]
\footnotesize
\caption{Less-forgetting for General Learning Cases}
\textbf{Input:} $\mathrm{D}_t, N_s, N_f, N_i, N_b$\\
\textbf{Output:} $\theta^{(t)}$\\
\begin{algorithmic}[1]
  \STATE $\theta^{(t)}\sim \mathcal{N}(0,\sigma^2)$
  \STATE for i=1,$\ldots$,$N_i$ // training iteration
  \STATE ~~~~if $i \mod N_s = 1$
  \STATE ~~~~~~~~$\theta^{(s)} \leftarrow  \theta^{(t)}$
  \STATE ~~~~end if
  \STATE ~~~~if $i \mod N_f = 1$
  \STATE ~~~~~~~~if $\lfloor{i / N_f}\rfloor=0$ or even number
  \STATE ~~~~~~~~~~~ Unfreeze the weights of the softmax layer of the target network.
  \STATE ~~~~~~~~else
  \STATE ~~~~~~~~~~~Freeze the weights of the softmax layer of the target network.
  \STATE ~~~~~~~~end if
  \STATE ~~~~end if
  \STATE ~~~~Select mini-batch set $\mathrm{B}$ from $\mathrm{D}_t$. ($|\mathrm{B}| = N_b$)
  \STATE ~~~~Update $\theta^{(t)}$ using backpropagation with $\mathrm{B}$ to minimize total loss $\mathcal{L}_t(x;\theta^{(s)}, \theta^{(t)})$ in Equation \ref{eq:total_loss}.
  \STATE end for
  \STATE Return $\theta^{(t)}$.
\end{algorithmic}
\label{alg:general}
\end{algorithm}

\section{Experiments}
\label{sec:exp}
We establish two different recognition experiments: unforgetting and generalization tests for Algorithm 1 and 2, respectively.
In the unforgetting experiment, we manually set up a source domain and a target domain using an object recognition dataset (CIFAR-10) and considered two different digit number recognition datasets (MNIST, SVHN) to belong to the source domain and the target domain, respectively. In the generalizaton experiment, we used CIFAR-10 dataset.
In addition, we used a dropout method \cite{srivastava2014dropout} for Maxout and LWTA experiments because we want to raise their maximum performance.


\subsection{Unforgetting Test}
In order to examine how much forgetting less prevents a DNN from losing its source domain information while also performing on the target domain, we manipulate CIFAR-10 whereas MNIST and SVHN remain intact. First, we split a training set of CIFAR-10 images, $\mathrm{D}_{t}$, into two sets so that they belong to the source and the target domains, $\mathrm{D}^{(s)}_{t},\mathrm{D}^{(t)}_{t}$; they contain 40,000 and 10,000 images, respectively. Second, we convert the pixel values of the images in $\mathrm{D}^{(t)}_{t}$ into new ones based on the equation $\frac{\mathrm{R}+\mathrm{G}+\mathrm{B}}{3}$, where $\mathrm{R}, \mathrm{G}, \mathrm{B}$ are the pixel values of each channel, and the same conversion is conducted on a test set, $\mathrm{D}_{v}$ where the converted set becomes $\mathrm{D}^{(t)}_{v}$.
However, in the digit unforgetting test, we use MNIST dataset to source domain, SVHN dataset to target domain.

\begin{figure}[h]
\begin{center}
\hspace{-3mm}
\includegraphics[width=4.3cm]{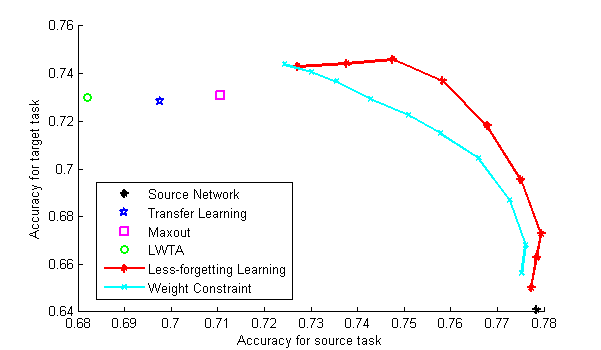}
\includegraphics[width=4.3cm]{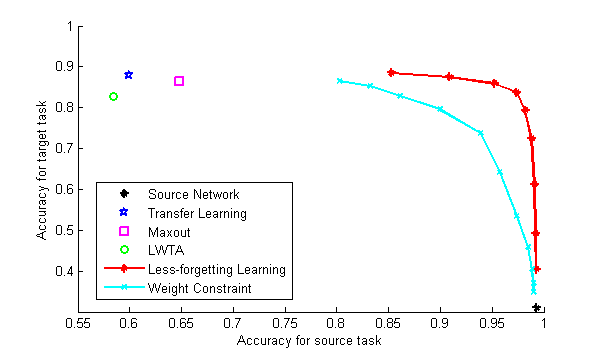}\\
\caption{\textbf{Source accuracy versus target accuracy.} On the left side is the object recognition source accuracy versus the target accuracy in CIFAR-10. On the right side is the digit recognition source (MNIST) accuracy versus the target (SVHN) accuracy. The accuracy curve is generated according to the value of $\lambda_e$ in Equation \ref{eq:total_loss}.}
\label{fig:roc}
\end{center}
\end{figure}

Table \ref{tb:Result} includes recognition rates of source networks, which are trained only on source domain and tested---with different activation functions---on each domain separately. The large accuracy gap between them certainly makes sense, since the DNNs has never seen examples from the target domain during training. 
As opposed to source network, we resume training with different methods on the target domain after learning is finished on the source domain nunder the constraint that during, target domain learning, DNNs cannot access examples from source domain. The results are also listed in Table \ref{tb:Result}. Two $\lambda_{e}$ values are selected based on observing Figure \ref{fig:roc}. Our method predicts true labels more accurately than previous works, such as the traditional transfer learning method, LWTA, and Maxout. In addition, this results provide compelling evidence that an alternating activation function is not the proper way to prevent a network from forget as little previously learned information as possible.

\begin{table}[t]
\renewcommand{\arraystretch}{1.2}
\centering
\caption{Experimental results.}
\vspace{-5mm}
\label{tb:Result}
\begin{center}
\begin{tabular}{c|ccc}
\hline
& & \bf{source} & \bf{target}\\ \hline
 & Source network (Maxout) \cite{goodfellow2013empirical} & 99.5 & 29.07 \\
&  Source network (LWTA) \cite{srivastava2013compete} & 99.5 & 27.5 \\
MNIST & Source network (ReLU) & 99.32 & 31.04 \\ \cline{2-4} 
$\downarrow$ & Transfer (ReLU)  & 59.93 & 87.83 \\ 
SVHN & Transfer (Maxout)  \cite{goodfellow2013empirical} & 64.82 & 86.44 \\
 & Transfer (LWTA)  \cite{srivastava2013compete} & 58.38 & 82.80 \\
 & LF ($\lambda_{e}=1.6*10^{-3}$) & \textbf{97.37} & \textbf{83.79} \\
 & LF ($\lambda_{e}=3.9*10^{-4}$) & \textbf{90.89} & \textbf{87.57} \\ \hline\hline
 & Source network (Maxout) \cite{goodfellow2013empirical} & 78.64 & 64.9 \\
& Source network (LWTA)  \cite{srivastava2013compete} & 76.04 & 65.72 \\
CIFAR10 & Source network (ReLU) & 77.84 & 64.09 \\ \cline{2-4} 
COLOR & Transfer (ReLU) & 69.4 & 70.84 \\ 
$\downarrow$ & Transfer (Maxout)  \cite{goodfellow2013empirical} & 71.06 & 73.07 \\
CIFAR10 & Transfer (LWTA)  \cite{srivastava2013compete} & 68.21 & 72.99 \\
GRAY & LF ($\lambda_{e}=1.6*10^{-3}$) & \textbf{75.83} & \textbf{73.70} \\ 
 & LF ($\lambda_{e}=3.9*10^{-4}$) & \textbf{73.77} & \textbf{74.43} \\ \hline
\end{tabular}
\end{center}
\vspace{-5mm}
\end{table}

Next, we observe the relation between target accuracy and domain accuracy in accordance with a value of $\lambda_e$, as displayed in Figure \ref{fig:roc}. It should be noted that the closer the curve comes to the top right corner, the better performance. Note that the top target recognition rate of less forgetting is even higher than that of transfer learning, LWTA, and Maxout in CIFAR-10. Therfore, it could be inferred that dissimilarity between source(color) and target domain(grayscale) is not that much, and remembering previous domain help more generlization.  

\subsection{Generalization Test}
As explained in Section \ref{sec:general}, we adopt Algorithm \ref{alg:general} on the object recognition and make a discovery that less forgetting learning shows better performance, i.e. more generalization. Table \ref{tb:general} shows the accuracy of original, batch normalization, batch normalization plus less forgetting, and less forgetting learning as the number of iteratioin increase. It is obvious that all cases equipped with less forgetting show an improvement over ones without it.

\begin{table}[h!]
\renewcommand{\arraystretch}{1.2}
\centering
\caption{Generalization Test.}
\vspace{-5mm}
\label{tb:general}
\begin{center}
\begin{tabular}{c||cc||cc}
\hline
&~~~Original~~~&~~~LF~~~&~~~BN~~~&~~~BN+LF~~~\\ \hline
Accuracy & 75.29 & \textbf{77.14} & 80.1 & \textbf{80.99}\\ \hline
\end{tabular}
\end{center}
\vspace{-5mm}
\end{table}

\section{Conclusion}
\label{sec:conclusion}
We proposed a less-forgetting learning method to alleviate a catastrophic forgetting problem in DNNs. Also, we observed that the forgetting phenomenon occurs when performing general learning method like a stochastic gradient descent method.
Our method is also effective to mitigate this problem. Finally, we showed that our method is useful for improving generalization ability of DNNs. 


%





\ifCLASSOPTIONcaptionsoff
  \newpage
\fi



%


{
\footnotesize
\bibliographystyle{IEEEtran}
\bibliography{dnn}
}

%








\end{document}